\documentclass{vgtc}

\usepackage{times}
\usepackage{enumitem}

\onlineid{0}

\vgtccategory{Research}

\vgtcinsertpkg

\usepackage[utf8]{inputenc}
\usepackage[english]{babel}
\usepackage{amsmath}
\usepackage{amsfonts}
\usepackage{amssymb}
\usepackage{graphicx}
\usepackage{bm} 

\usepackage[dvipsnames]{xcolor}

\usepackage{caption}
\usepackage{subfig}

\usepackage[numbers]{natbib} 
\usepackage[hidelinks]{hyperref} 

\usepackage{wrapfig} 

\usepackage{float}
\floatstyle{plaintop}
\restylefloat{table}

\usepackage{natbib}

\usepackage{lineno}
\modulolinenumbers[5]

\title{Real-time texturing for 6D object instance detection from RGB Images}

\author{Pavel Rojtberg \thanks{pavel.rojtberg@igd.fraunhofer.de} \\
\parbox{1.4in}{\scriptsize \centering Fraunhofer IGD, Darmstadt \\ TU Darmstadt}
\and
Arjan Kuijper \thanks{arjan.kuijper@igd.fraunhofer.de} \\
\parbox{1.4in}{\scriptsize \centering Fraunhofer IGD, Darmstadt \\ TU Darmstadt}
}

\makeatletter

\hypersetup{
pdftitle={\@title}
}

\CCScatlist{ 
  \CCScat{I.2.10}{ARTIFICIAL INTELLIGENCE}{Vision and Scene Understanding}{Modeling and recovery of physical attributes};
	\CCScat{I.5.5}{PATTERN RECOGNITION}{Implementation}{Interactive systems};
}

\abstract{For objected detection, the availability of color cues strongly influences detection rates and is even a prerequisite for many methods. However, when training on synthetic CAD data, this information is not available.
We therefore present a method for generating a texture-map from image sequences in real-time. The method relies on 6 degree-of-freedom poses and a 3D-model being available. In contrast to previous works this allows interleaving detection and texturing for upgrading the detector on-the-fly.
Our evaluation shows that the acquired texture-map significantly improves detection rates using the LINEMOD \citep{hinterstoisser2012model} detector on RGB images only.
Additionally, we use the texture-map to differentiate instances of the same object by surface color.
}

\begin{document}

\firstsection{Introduction}

\maketitle

In recent years there has been great progress on the task of object detection and 6D pose estimation by means of convolutional neural networks (CNN) \cite{tekin2018real, kehl2017ssd}.
Combined with a successive local refinement method \cite{manhardt2018deep, Seo13}, it is now possible to obtain a precise object pose from a single RGB image only.

However, these methods take advantage of 3D scans of the target objects to generate training data. The 3D scans not only provide geometry but also surface information.
Although scans are reasonably easy to acquire \cite{newcombe2011kinectfusion}, the need of a 3D scan restricts the availability of the methods.

For many use-cases only a model created with computer aided design (CAD) software is available. Such models are coloured by semantics (e.g. engine, wheel) instead of material appearance (e.g. metal, rubber).
Therefore, when using CAD reference geometry for object detection, one can not rely on surface colour or texture.
This scenario is common in industrial environments or with 3D printing, where a manifold of materials can be used to represent the same CAD geometry.

In these cases depth-only variants of some algorithms \cite{tekin2018real, kehl2017ssd}  can be used --- however at the cost of degraded performance.

\begin{figure}
\includegraphics[width=0.24\textwidth]{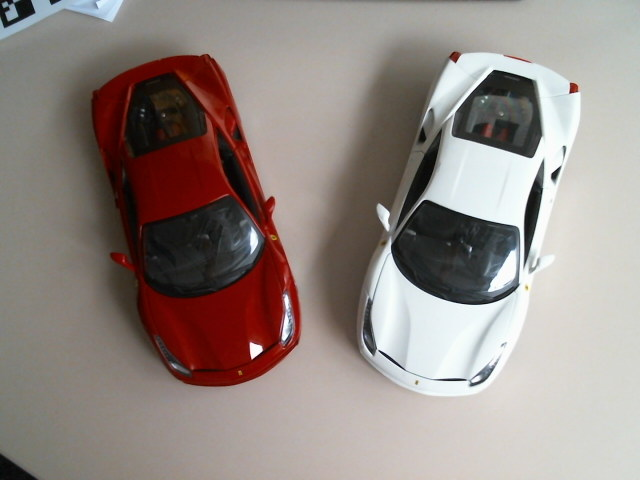}
\includegraphics[width=0.24\textwidth]{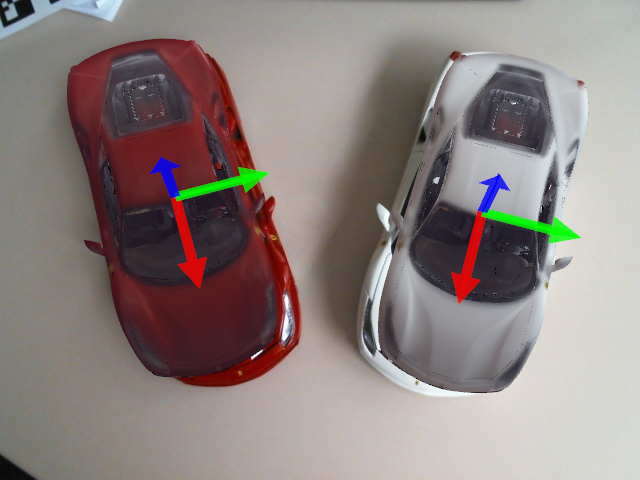}
\caption{Our extension of LINEMOD \citep{hinterstoisser2012model} is able to correctly detect multiple instances of the same object based on surface color. The corresponding 6D poses are visualized using the textured mesh.}
\label{fig:instances}
\end{figure}

This work therefore focuses on the real-time acquisition of surface texture data and dynamic detection-model augmentation. This allows capturing and using the surface color information on-the-fly.

In the context of real-time surface reconstruction there is notably the work by \citet{whelan2013robust}, who extends the Kinect Fusion \cite{newcombe2011kinectfusion} algorithm to colors. For this, an additional 3D color volume of the same size as the geometry voxel grid is used. This means that the surface resolution is tied to the geometry resolution and the memory consumption has cubic complexity. Furthermore, RGB-D data is required. The mentioned 3D scans are typically acquired by variations of \cite{newcombe2011kinectfusion} and consequently store surface information as vertex colors. 
Here, \cite{zhou2014color} improve surface resolution by subdividing the scanned geometry. This is similar to using texture maps but results in inhomogeneous sampling and inefficient storage.

On the other hand, in the context of 3D scanning \cite{callieri2008masked}, 2D textures are often used which only have quadratic storage complexity and allow decoupling surface resolution from geometric resolution. More specifically, structure-from-motion based methods \cite{waechter2014let} only require RGB images to reconstruct both surface and geometry.

However, these methods operate on a-priori recorded data-sets which prevents real-time operation. Notably, the global optimization step alone, as employed by those methods, takes up to several hours.
Our method in contrast operates in real-time while only requring RGB frames to incrementally generate a 2D texture. To this end, we assume all geometry as fixed and given and only optimize locally for color consistency.
The closest method to our work is by \citet{magnenat2015live}, who also map a 2D camera image to texture in real-time. However, their work specifically only addresses a single view and focuses on the in-painting aspect.

To employ our texturing method for object detection, we build upon the LINEMOD detection framework by \cite{hinterstoisser2012model}.
They employ a two-stage, handcrafted feature descriptor, specifically tuned for texture-less object detection. In the first step gradient templates (DOT) are matched to the input image in a sliding window fashion, which capture the contour of an object. In a successive outlier-rejection step, surface color is used to filter implausible matches, based on the interior color.

Even though this no longer provides state-of-the-art detection performance \cite{tekin2018real, kehl2017ssd}, the internal separation allows computing the DOT features on CAD geometry only and add the surface color at run-time.
This is generally not possible with deep learning based approaches, which rely on surface color being available during training. Efforts to train on an abstract representation \cite{rambach2018learning} to allow for different object appearances, typically result in a degraded performance compared to training on real images.
In contrast, our extension of \cite{hinterstoisser2012model} improves its performance, while allowing to differentiate several instances of the same geometric object by their surface properties.

Based on the above, our key contributions are;
\begin{enumerate}
\item an incremental, real-time texture-map extraction pipeline and
\item efficient integration of texture-maps for object instance recognition.
\end{enumerate}

This paper is structured as follows: in Section \ref{sec:uvmap} we present our texture extraction algorithm in detail. Section \ref{sec:recognition} then describes the application of the texture-maps for object instance recognition, while in Section \ref{sec:evaluation} the use of texture-maps for object detection is evaluated using the LINEMOD dataset \cite{hinterstoisser2012model}.

We conclude with Section \ref{sec:conclusion} giving a summary of our results and discussing the limitations and future work.


\section{Texture extraction}
\label{sec:uvmap}
\begin{figure}
\subfloat[Image space] {
\includegraphics[width=0.24\textwidth]{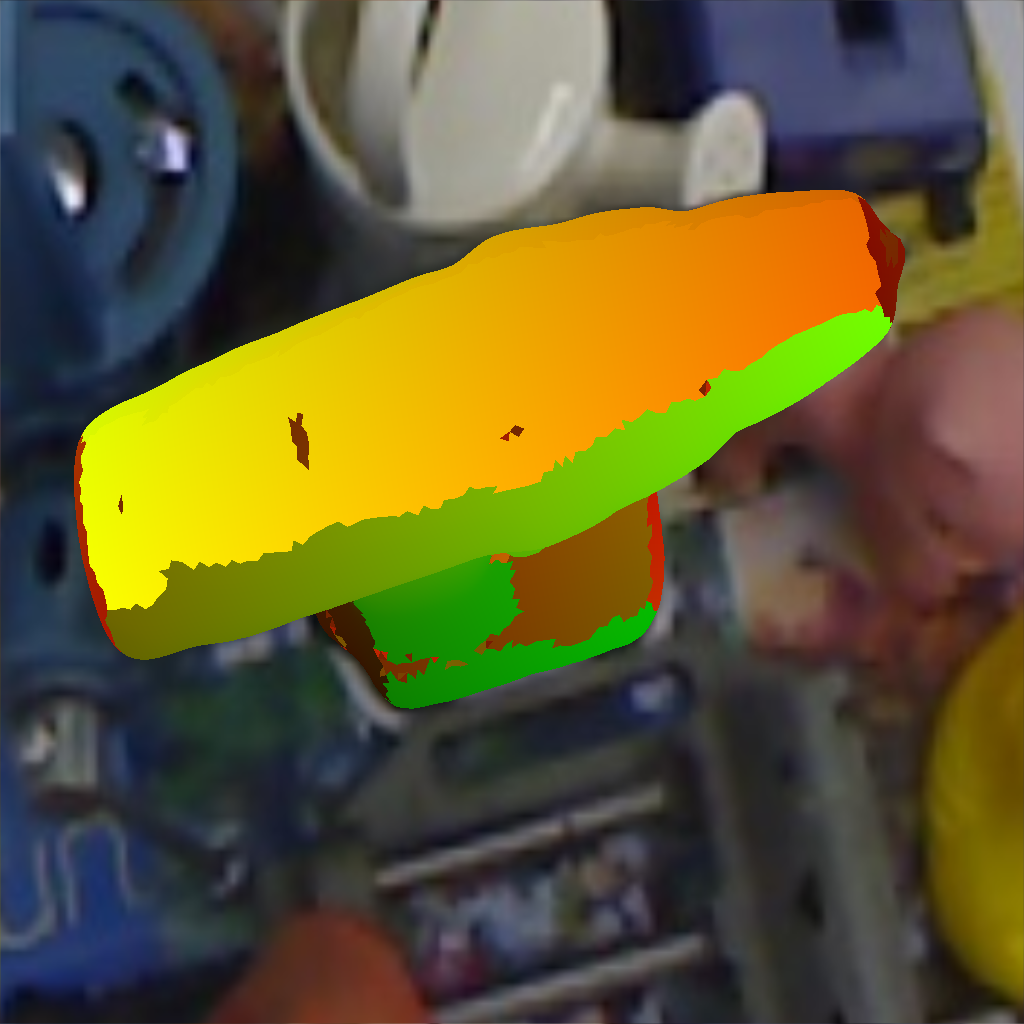}
\label{fig:imageuv}
}
\subfloat[Texture space] {
\includegraphics[width=0.24\textwidth]{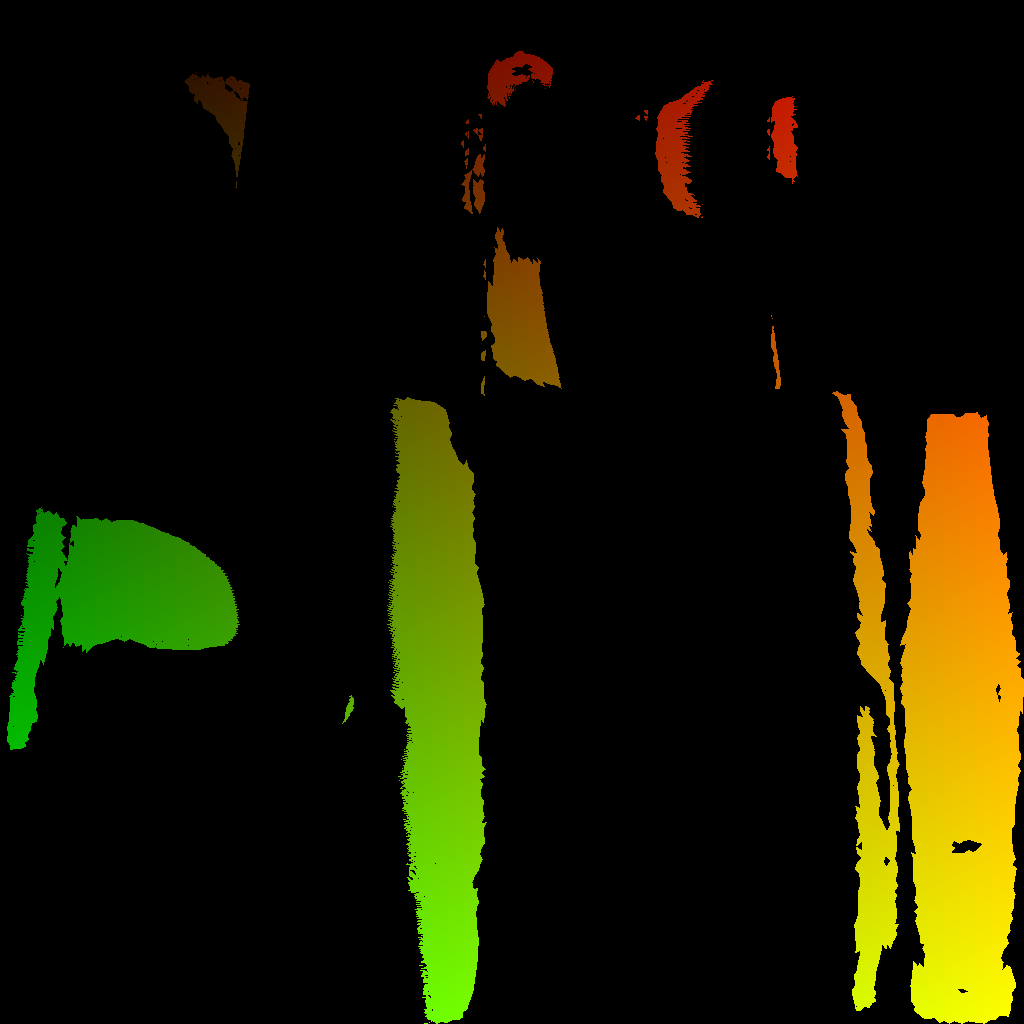}
\label{fig:depth_test}
}
\caption{We are mapping from image to texture space, which is the reverse direction compared to rendering. Texture coordinates are encoded as red-green.}
\label{fig:mapping}
\end{figure}

In rendering, the process of "texture mapping" consist of the following two steps
\begin{enumerate}
\item creating a mapping from a texture to the surface of a 3D model and
\item projecting the model and simultaneously mapping the texture into a 2D image.
\end{enumerate}
The first step is also called "texture atlas creation" and is typically performed by an artist during mesh creation. 
In contrast, our focus is the reverse direction, namely mapping from a 2D image of a projected 3D model back to the surface image as specified by the texture atlas (see Figure \ref{fig:mapping}).

Generally texture atlas coordinates are not included in CAD data and therefore have to be generated. However, automatic texture atlas generation is still an active area of research \cite{li2018optcuts} and outside of the scope of this work.
Here, we just use the angle-based "Smart UV Project" algorithm implemented in the Blender toolset (v2.79b) to generate the texture atlas and instead focus on the second step of texture mapping.

\begin{figure*}
\includegraphics[width=1.0\textwidth]{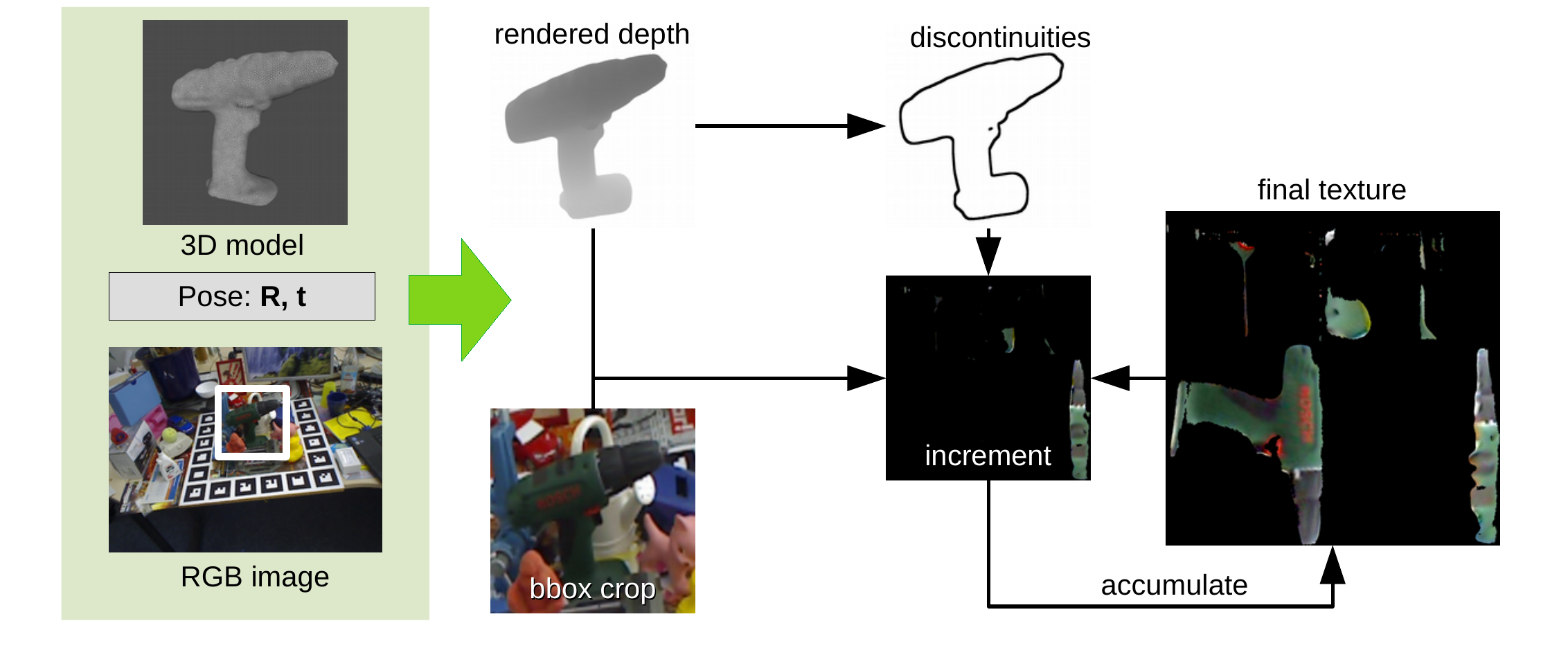}
\caption{Our texture extraction pipeline. Given a 3D model and its pose in a RGB frame, we first render the depth to determine visibility. Image regions around depth discontinuities are discarded as they are unreliable.
Next an texture-increment is extracted and a per-pixel score is computed to decide whether to merge the visible pixels into the final texture. Only the following buffers are required on the GPU; "final texture", "increment" and "discontinuities".}
\label{fig:pipeline}
\end{figure*}

In the remainder of this section we first discuss a simple exposure normalization scheme, before we present our texture extraction method in detail and finally turn to merging multiple views into one texture. The full pipeline is illustrated in Figure \ref{fig:pipeline}.

\subsection{Exposure normalization}
As our method does not explicitly compensate for different exposure times we pre-process the image stream to homogenize the brightness.
For this we use the first captured frame as reference and modify the successive frames to match its brightness and contrast levels.

Here we follow the idea of \citet{reinhard2001color} of adapting an input image $\textbf{I}$ to match a reference image as

\begin{equation}
\textbf{I}_n = \dfrac{\sigma_{ref}}{\sigma_{I}} \cdot \left( \textbf{I} - \mu_{I} \right) + \mu_{ref}
\end{equation}

where $\mu_I, \sigma_I$ and $\mu_{ref}, \sigma_{ref}$ are the mean and variance of the input image and the reference image, respectively.

However, whereas \cite{reinhard2001color} apply the transfer for all channels in the Lab color space, we only apply it to the luma component Y in the YUV color space as
we explicitly want to preserve the chrominance information.

This step is omitted if the exposure can be fixed during capturing.

\subsection{Texture-space to image-space mapping}

Texture mapping can be formalized as follows: given a triangulated mesh, each vertex $\mathbf{v}_i = [X, Y, Z, 1]$ with an associated texture coordinate $\mathbf{t}_i = [u, v]$ is projected into the current view by a world-to-image transform $\mathbf{P}$ as $\mathbf{p}_i = \mathbf{P} \cdot \mathbf{v}_i$. Here $\mathbf{p} = [x, y, 1]$ is a normalized pixel location in the image $\mathbf{I}$.

On the interior of the triangle formed by $(\mathbf{t}_i, \mathbf{t}_j, \mathbf{t}_k)$, a texture coordinate $\hat{\mathbf{t}}$ is interpolated and used for lookup in texture $\mathbf{T}$ as
\begin{equation}
\mathbf{I}(\mathbf{p}) = \mathbf{T}(\hat{\mathbf{t}}).
\label{eq:texmap}
\end{equation}
This mapping is continuous in texture space and therefore allows for bi-linear interpolation to avoid aliasing artifacts.

For texture extraction however we are interested in the reverse mapping, namely
\begin{equation}
\mathbf{T}(\mathbf{t}) = \mathbf{I}(\hat{\mathbf{p}}).
\end{equation}
Instead of iterating over the mesh topology as defined by $\mathbf{v}_i$ in 3D, we now iterate over $\mathbf{t}_i$ as defined by the texture-atlas in 2D. Conversely, we now require a continuous value of $\hat{\mathbf{p}}$ in image space for lookup. This is computed by interpolating in the triangle formed by $(\mathbf{p}_i, \mathbf{p}_j, \mathbf{p}_k)$, of which each point is obtained as above by $\mathbf{p}_i = \mathbf{P} \cdot \mathbf{v}_i$.

Here, visibility must be explicitly computed; with equation \eqref{eq:texmap}, we implicitly assumed overlapping points to be resolved by a depth-test, only retaining the points closest to the camera. This can no longer be exploited, as points do not overlap in the texture space.

\begin{figure}
\subfloat[Depth test aliasing] {
\includegraphics[width=0.24\textwidth]{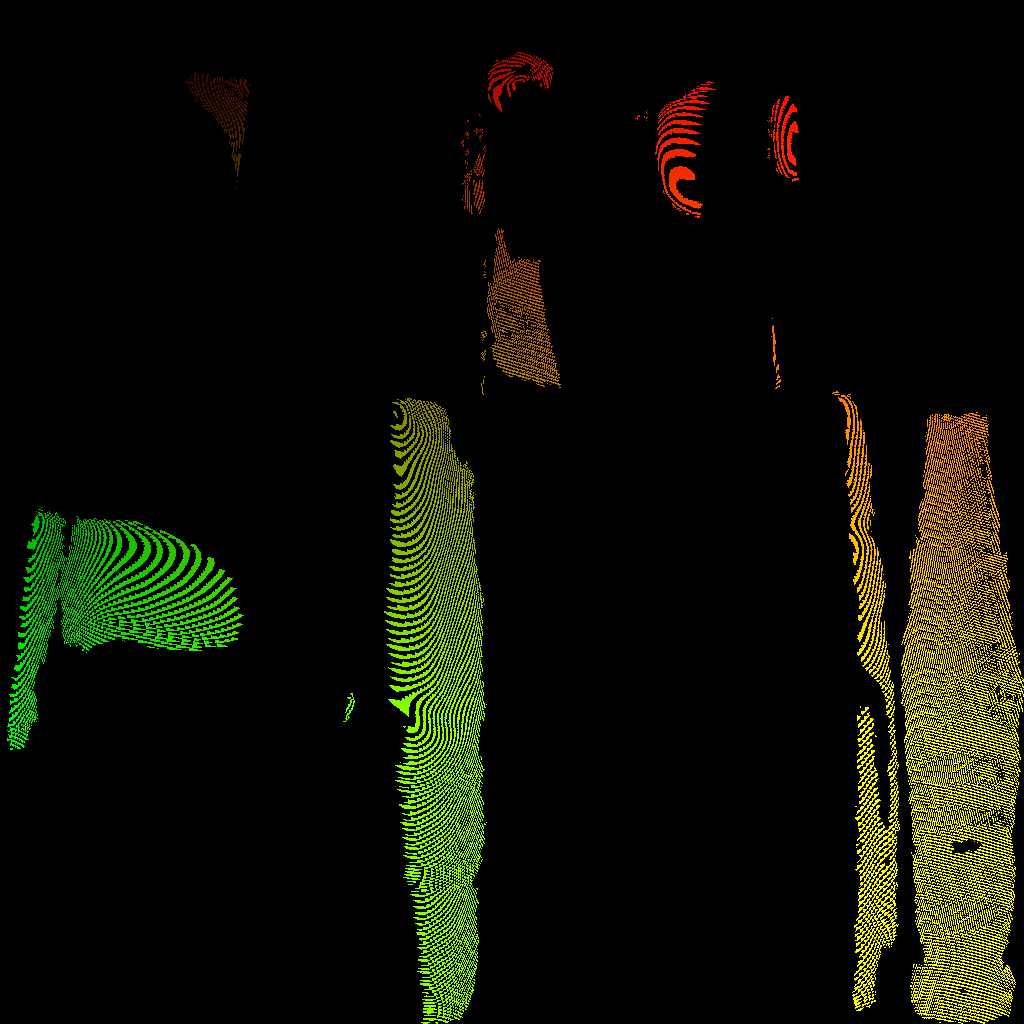}
\label{fig:depth_alias}
}
\subfloat[Slope biased depth] {
\includegraphics[width=0.24\textwidth]{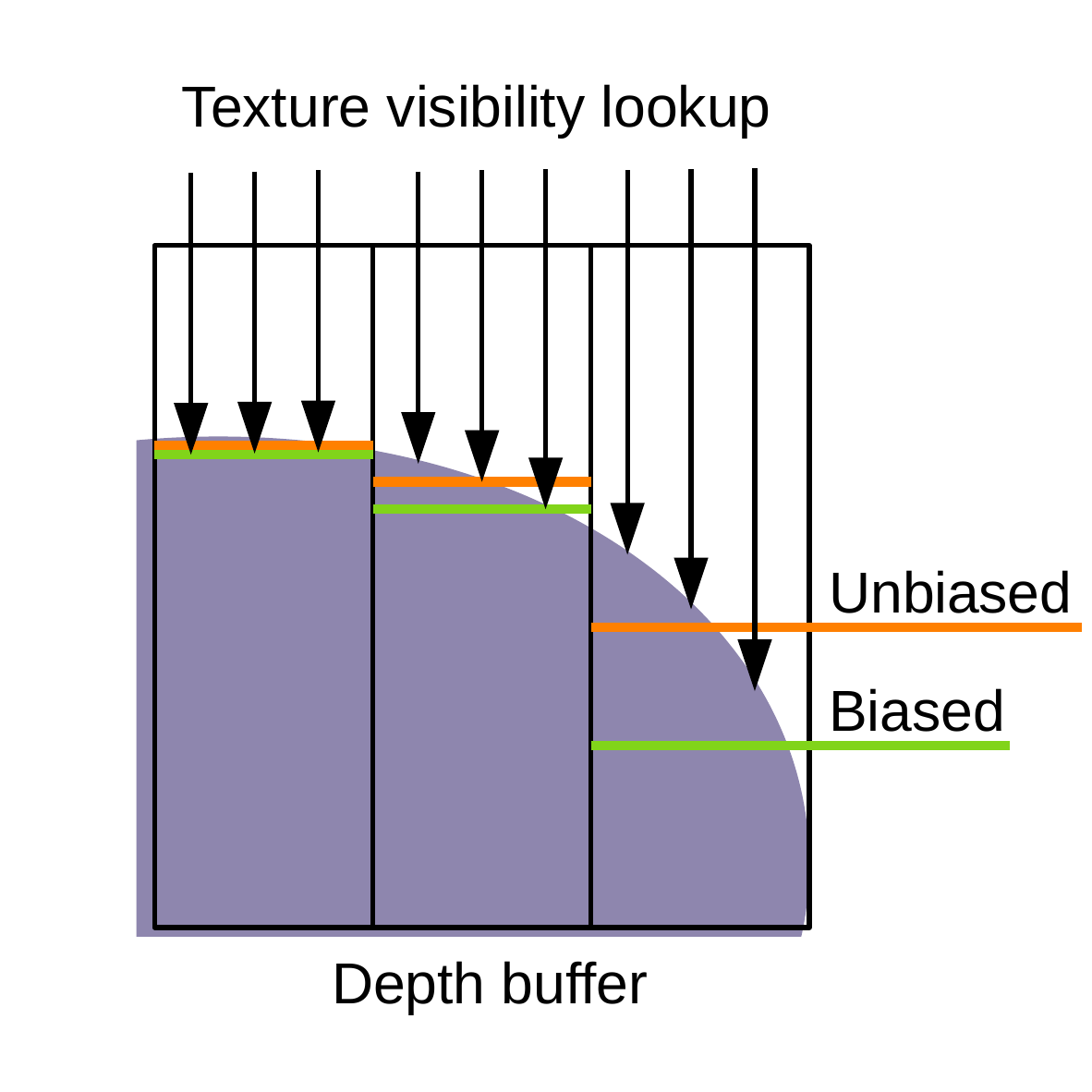}
\label{fig:depth_bias}
}
\caption{We store slope-scaled biased depth values to avoid aliasing errors during the visibility test.}
\end{figure}

To handle visibility we therefore introduce an additional depth buffer and render depth from the camera view. This allows comparing the depth of an interpolated coordinate $\hat{\mathbf{p}}$ to the actually visible depth value.
However, this leads to aliasing; with non-planar objects the view resolution cannot be adapted to match the texture space resolution.

To remedy the aliasing artifacts we apply techniques from the shadow mapping domain \cite{brabec2002practical} where the same problem occurs when a scene is rendered from a shadow camera and an observer camera view. Particularly, we

\begin{enumerate}
\item focus the camera on the object bounding box to increase the sampling rate in image space and
\item apply a slope-scale depth-bias to account for the remaining differences in sampling rates during visibility testing.
\end{enumerate}
The latter is especially important; as the texture atlas has a higher sampling rate than the depth buffer, several points $\hat{\mathbf{p}}$ , interpolated in the texture space, map to the same point $\mathbf{p}$ in the image depth-buffer. 
At steep angles $\hat{\mathbf{p}}$ has a strong depth variation and thus neighboring points alternatively fail and pass the visibility test when compared to a single reference value (see Figure \ref{fig:depth_bias}).

To account for this we store a biased depth that allows for a sampling offset of 1px in image space. The bias $b$ depends on the depth slope $dz$ per pixel $dx$ and the minimal depth buffer resolution $r$ as:
\begin{equation}
b = \dfrac{dz}{dx} + r.
\end{equation}
The bias is large in steep regions while minimal for faces parallel to the camera. This computation can be implemented efficiently on the GPU by using e.g. \textit{glPolygonOffset}.
The effect can be observed by comparing Figure \ref{fig:depth_alias} and Figure \ref{fig:depth_test}.
This allows us to map each visible pixel from a single image into the texture to record the object surface.
The resulting reconstruction can already be applied for detecting the object in similar views (see Section \ref{sec:recognition}).

\subsection{Merging multiple views}

Generally the object surface is only partially visible from a single view and therefore multiple images are needed to reconstruct the full texture.

\begin{figure}
\subfloat[Initial view] {
\includegraphics[width=0.24\textwidth]{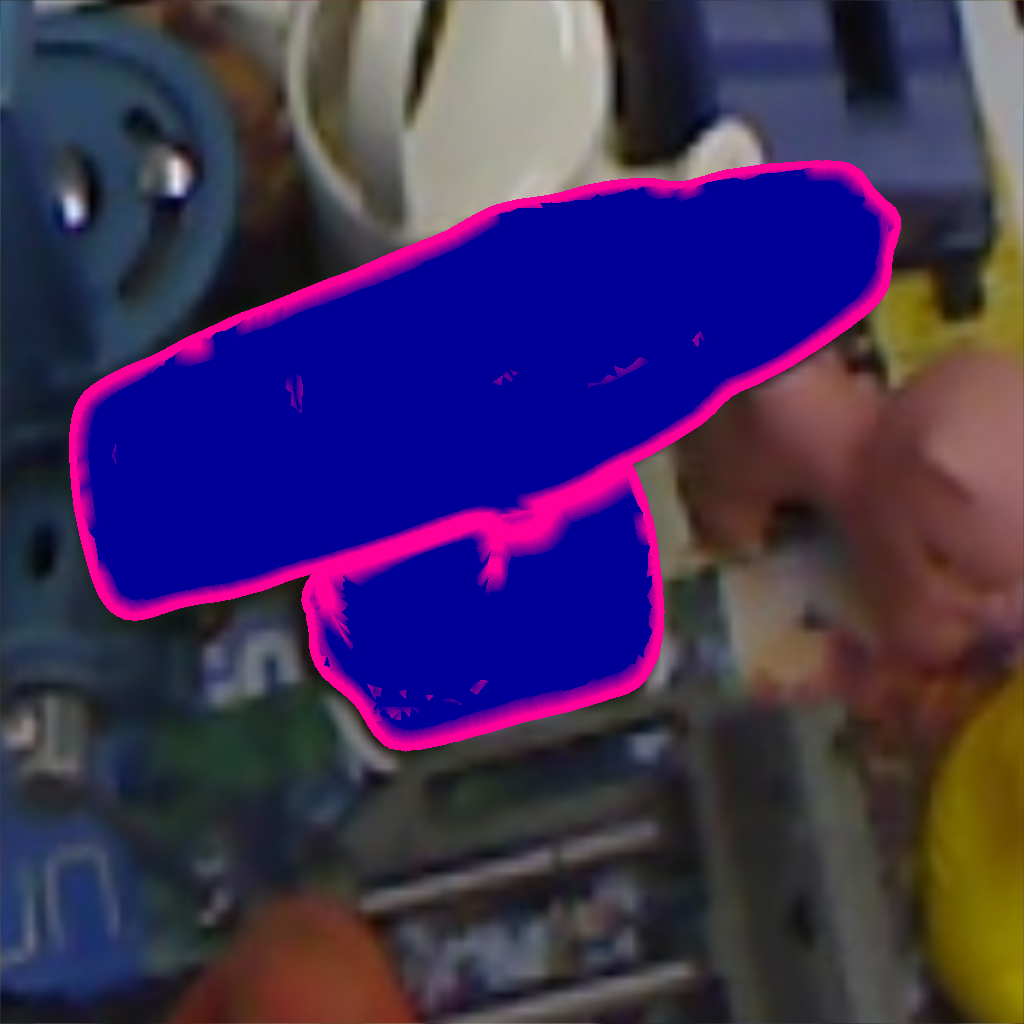}
\label{fig:initmerge}
}
\subfloat[argmax based update with blending] {
\includegraphics[width=0.24\textwidth]{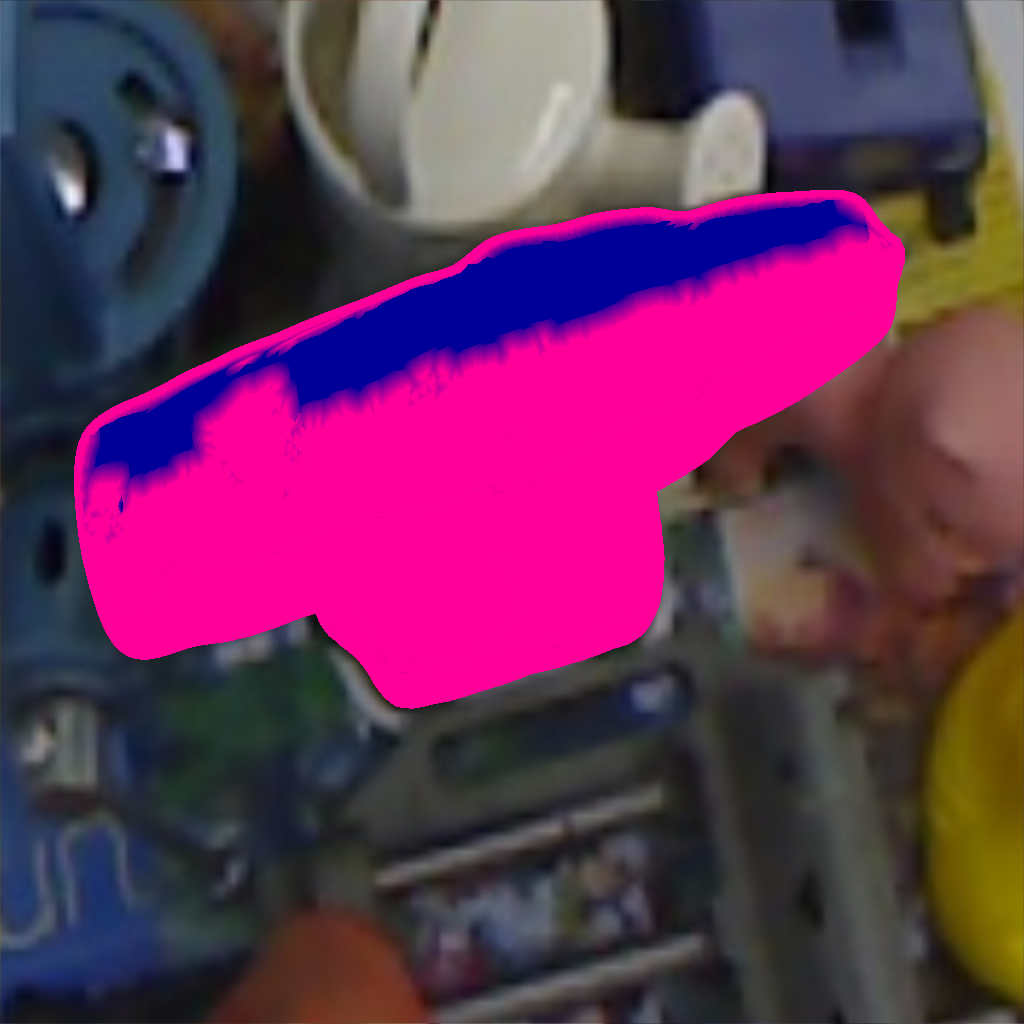}
\label{fig:mergemap}
}
\caption{Merge-maps of two successive frames when using Eq. \eqref{eq:max}. Valid pixels are colored blue.}
\end{figure}

Assuming that the same texture point $\mathbf{t}$ will be observed in different images as $\mathbf{c_0}, \ldots, \mathbf{c_N}$ where $\mathbf{c_i} = \mathbf{I}_i(\hat{\mathbf{p}}_i)$, we discard edge-pixels at object boundaries or strong depth discontinuities. These measurements are unreliable as they might come from different surfaces due to pose imprecisions and limited camera resolution. Instead, we aim for a view where $\mathbf{t}$ is not at an observed edge. A pixel is considered to be part of an edge if the depth change is larger than 10\% of the object diameter.
All points in a 5px neighborhood of an edge-pixel are discarded as well (see Figure \ref{fig:initmerge}).

To combine multiple valid observations $\mathbf{c_i}$ of $\mathbf{t}$, we define score $s$ that, inspired by \cite{callieri2008masked}, weighs each observation by the distance to camera $d$ and the angle $\alpha$ between surface normal and view direction as
\begin{equation}
s = \operatorname{cos} \alpha \cdot (1 - d),
\end{equation}
where $d$ is assumed in normalized device coordinates ranged $[0; 1]$ and $\alpha$ is computed based on the interpolated surface normal, which can be defined per vertex $\mathbf{v}_i$ (e.g. for a sphere)
and therefore is not required to be constant for a single face.

Using $s$ we implemented two merging strategies; a weighted arithmetic mean

\begin{equation}
\mathbf{t} =  \dfrac{s_0 \cdot \mathbf{c_0} + \ldots + s_n \cdot \mathbf{c_n}}{s_0 + \ldots + s_n},
\label{eq:mix}
\end{equation}
and only retaining the best view
\begin{equation}
\mathbf{t} = \underset{\mathbf{c_i}}{\operatorname{arg\,max}} \, \left\lbrace s_0, \ldots, s_n \right\rbrace.
\label{eq:max}
\end{equation}
Both equations can be efficiently implemented on the GPU using a single RGBA buffer for accumulation, as $RGBA = [\mathbf{c}, s]$.

\begin{figure*}
\subfloat[Vertex colors] {
\includegraphics[width=0.24\textwidth]{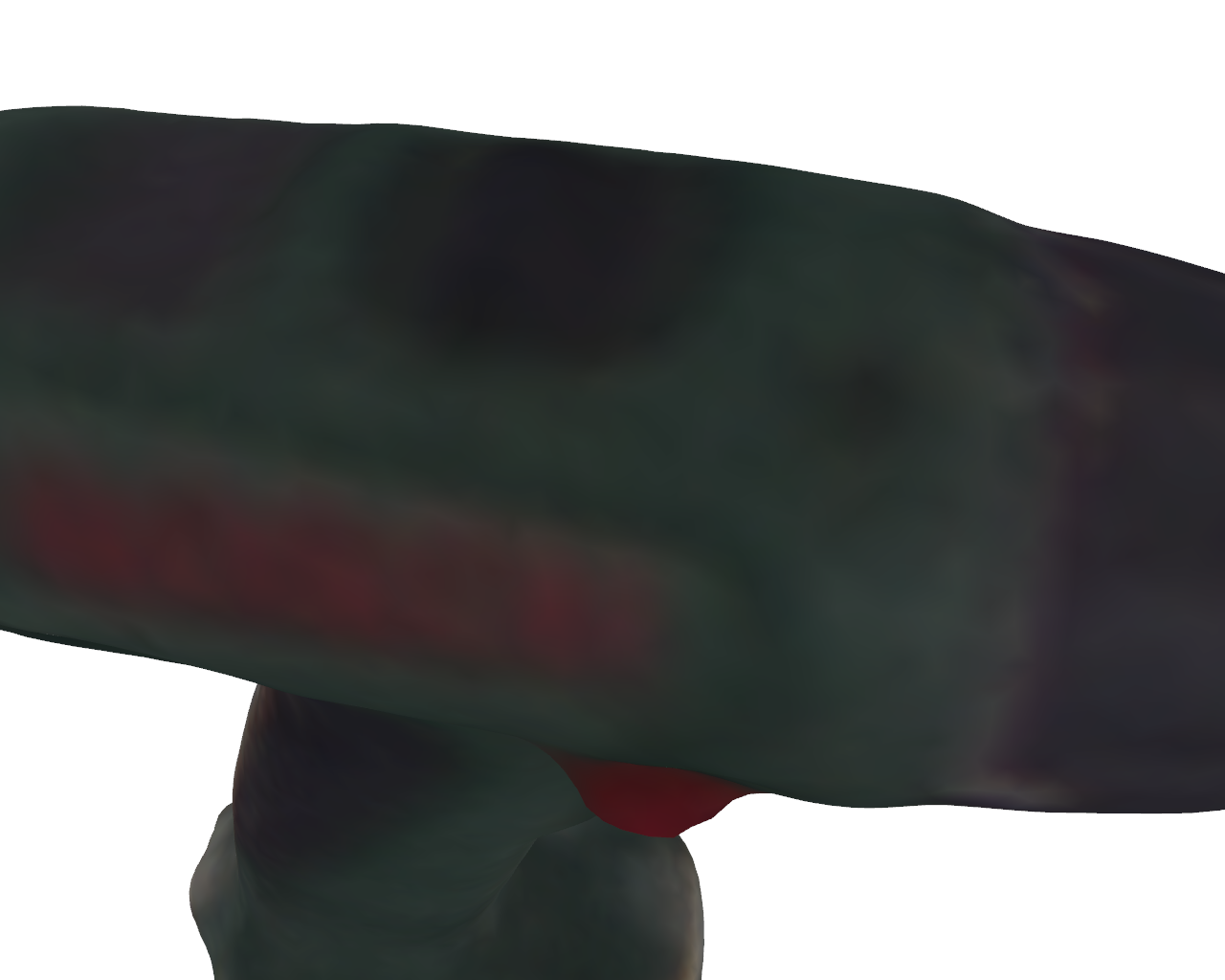}
}
\subfloat[Weighted mean as in Eq. \eqref{eq:mix}] {
\includegraphics[width=0.24\textwidth]{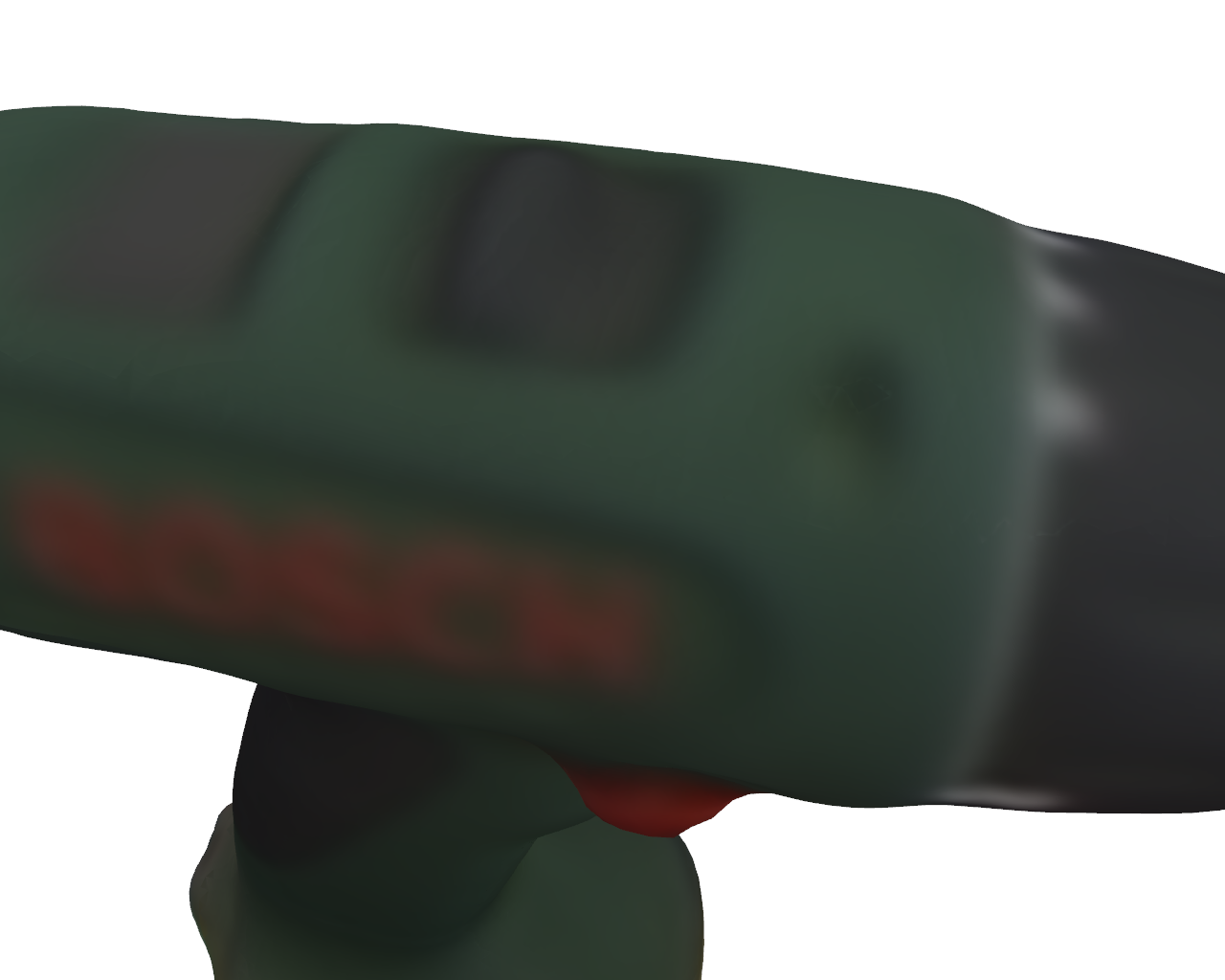}
\label{fig:texmix}
}
\subfloat[Best score as in Eq. \eqref{eq:max}] {
\includegraphics[width=0.24\textwidth]{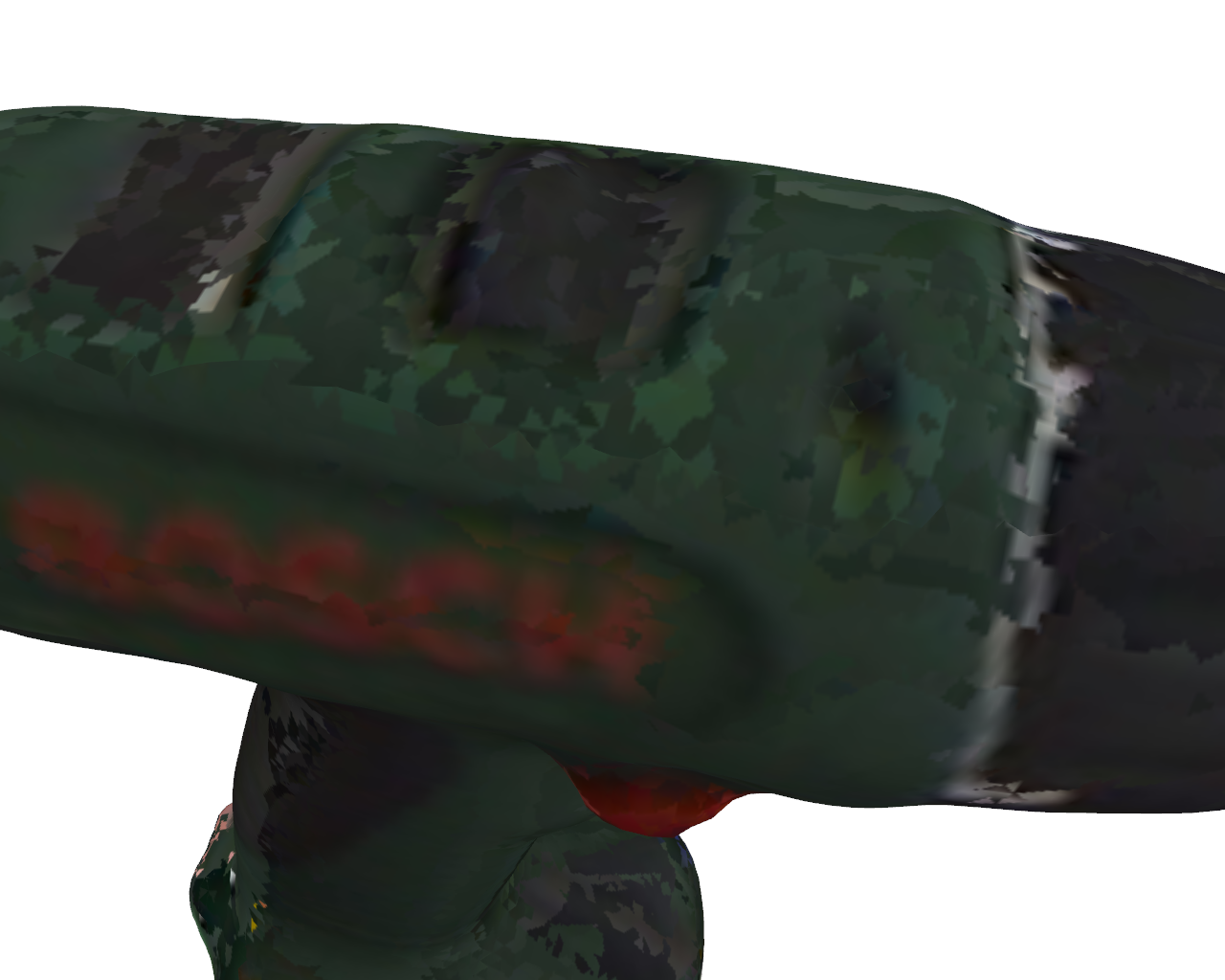}
\label{fig:texmaxnoblend}
}
\subfloat[Best score + blending] {
\includegraphics[width=0.24\textwidth]{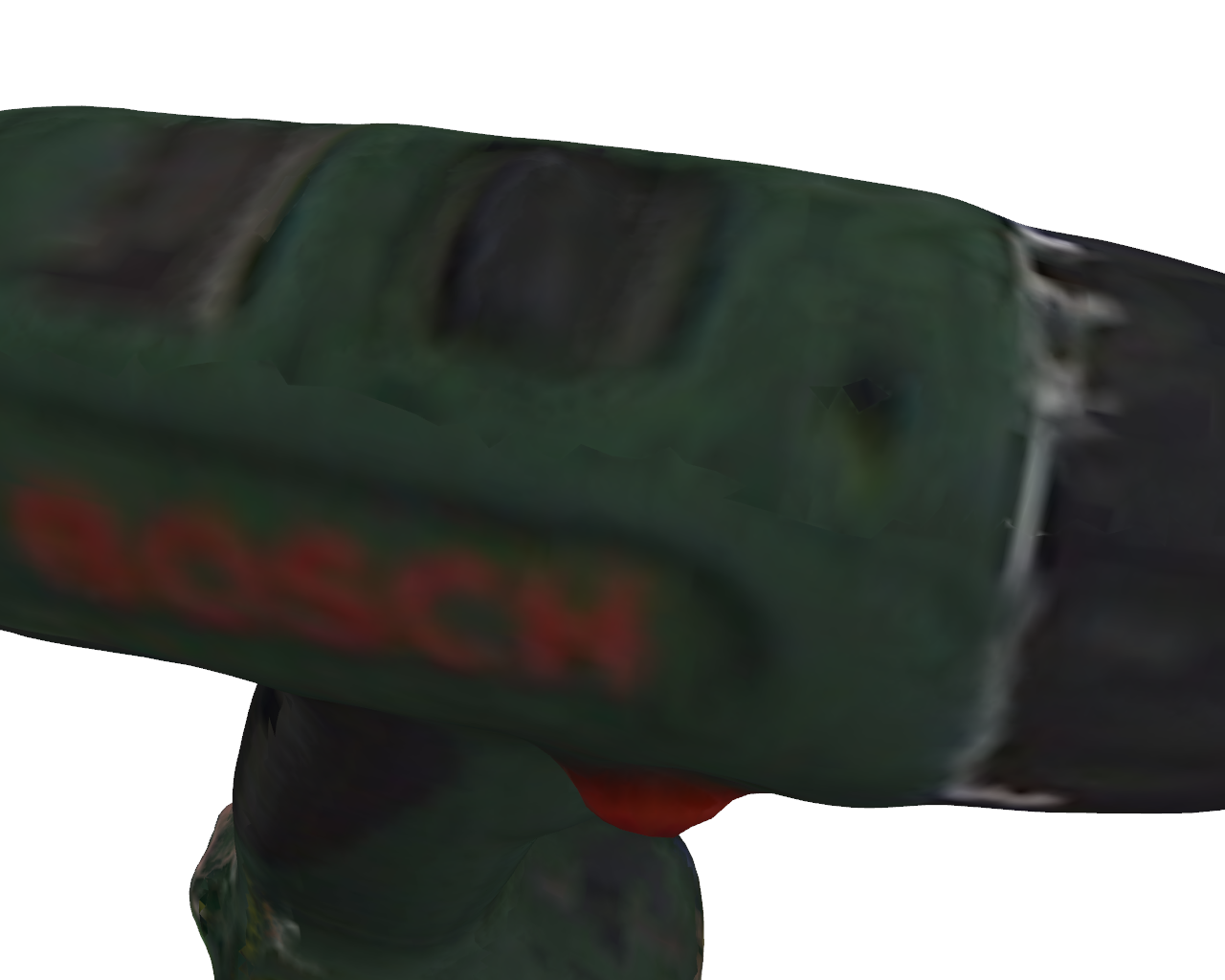}
\label{fig:texmax}
}
\caption{Exemplary surface color reconstructions of the "Driller" object Texture merging strategies using (a) KinectFusion and (b, c, d) variations of our algorithm.}
\label{fig:surface_reconstructions}
\end{figure*}

Figure \ref{fig:surface_reconstructions} shows exemplary results. Eq. \eqref{eq:mix} produces a smooth surface, while retaining more detail than vertex coloring. However, the averaging over slightly inaccurate object poses results in a loss of fine detail when compared to Eq. \eqref{eq:max}.

Using Eq. \eqref{eq:max} on the other hand retains all details, but emphasizes inconsistencies in exposure or object pose as seams between neighboring increment texture-patches.

To alleviate this problem we blend increment-patches at their boundaries into the existing texture during accumulation. Instead of simply overwriting the texture content with the new maximum, we compute the distance transform to the patch boundaries over a 5x5px support using the L2 norm. Using the distance we then linearly interpolate between the old and the new color value $\mathbf{c}$ and pixel score $s$.

Figure \ref{fig:mergemap} shows an increment-patch for Figure \ref{fig:initmerge}, projected onto the object. Note the gradient at the edges, which is linear in texture space.

The blending not only produces visually more pleasing results (compare Figures \ref{fig:texmaxnoblend} and \ref{fig:texmax}), but is crucial for computing the LINEMOD descriptor which relies on local gradient orientation.

\section{Object instance detection}
\label{sec:recognition}

In this section we describe how to employ the extracted textures for object instance detection i.e. differentiating multiple instances of the same object.
Here, we extend the color based outlier rejection of \cite{hinterstoisser2012model} to multiple color hypotheses to simultaneously perform classification.

The idea of color based outlier rejection in \cite{hinterstoisser2012model} is to store the expected color of the object projection alongside the LINEMOD template and at run-time count how many pixels in the camera frame have the expected color.

\begin{figure}
\subfloat[input image] {
\includegraphics[width=0.195\textwidth]{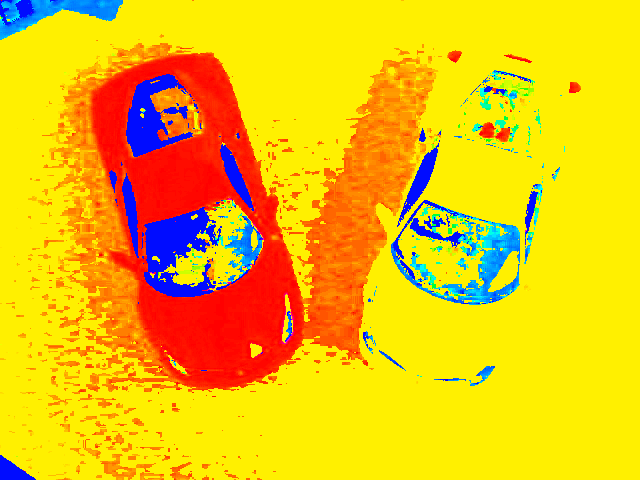}
\label{fig:hue_descr}
}
\subfloat[candidate / white template / red template] {
\includegraphics[width=0.2835\textwidth]{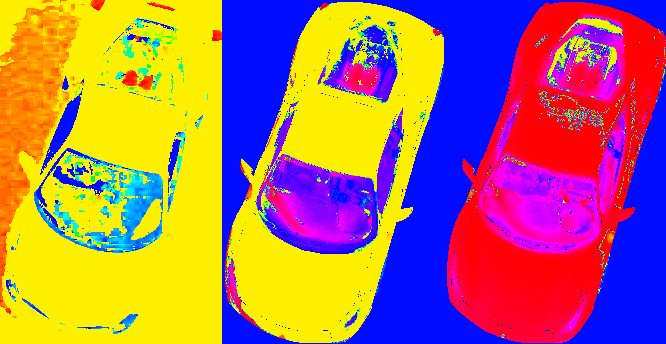}
\label{fig:huehyp}
}
\caption{Hue based instance detection. The input image (Figure \ref{fig:instances}) is cropped based on the template bounding box and compared to a set of hue templates.}
\end{figure}

To make the check robust against lighting variations, they convert the images to the HSV colour space and compare only the hue component. However, hue does not cover the colors black ($V=0$) and white ($V=1, S=0$).
Therefore, these are mapped to blue and yellow respectively, which completes the color based descriptor (see Figure \ref{fig:hue_descr}).

To extend this scheme for object instance detection as well as for on-the-fly recorded textures, we separate the expected color from the expected surface visibility. To this end, we store the texture coordinates of the object projection (compare Figure \ref{fig:imageuv}) instead of storing the expected color directly. The template surface-texture is stored separately.
At runtime we now use the texture coordinates to perform a lookup into the template-texture to retrieve the expected color, which gives us the same information as in \cite{hinterstoisser2012model}.

However, it is now possible to easily swap the surface-texture to globally change the expected colors.
Here a live-reconstructed texture can provide more accurate template colors and notably multiple template-textures can be used for object instance detection (see Figure \ref{fig:huehyp}).

Finally, the outlier rejection scheme needs a slight modification for classification. Instead of returning the first inlier based on the expected color, it needs to allow multiple matches without repetition.
For this, after finding an inlier, only the corresponding texture-template is removed and the remaining candidates are checked until all template-textures are found or all candidates are rejected.

While this is an integral part of the LINEMOD pipeline, it can be optionally integrated as a post-processing step to an CNN based architecture that is capable to abstract the object appearance to some degree.
E.g. it can be exectued after non-maximum-suppression in \cite{tekin2018real} to compute agreement with the color template.

\section{Evaluation}
\label{sec:evaluation}

\begin{figure*}
\subfloat[LINEMOD \cite{hinterstoisser2012model}] {
\includegraphics[width=0.3\textwidth]{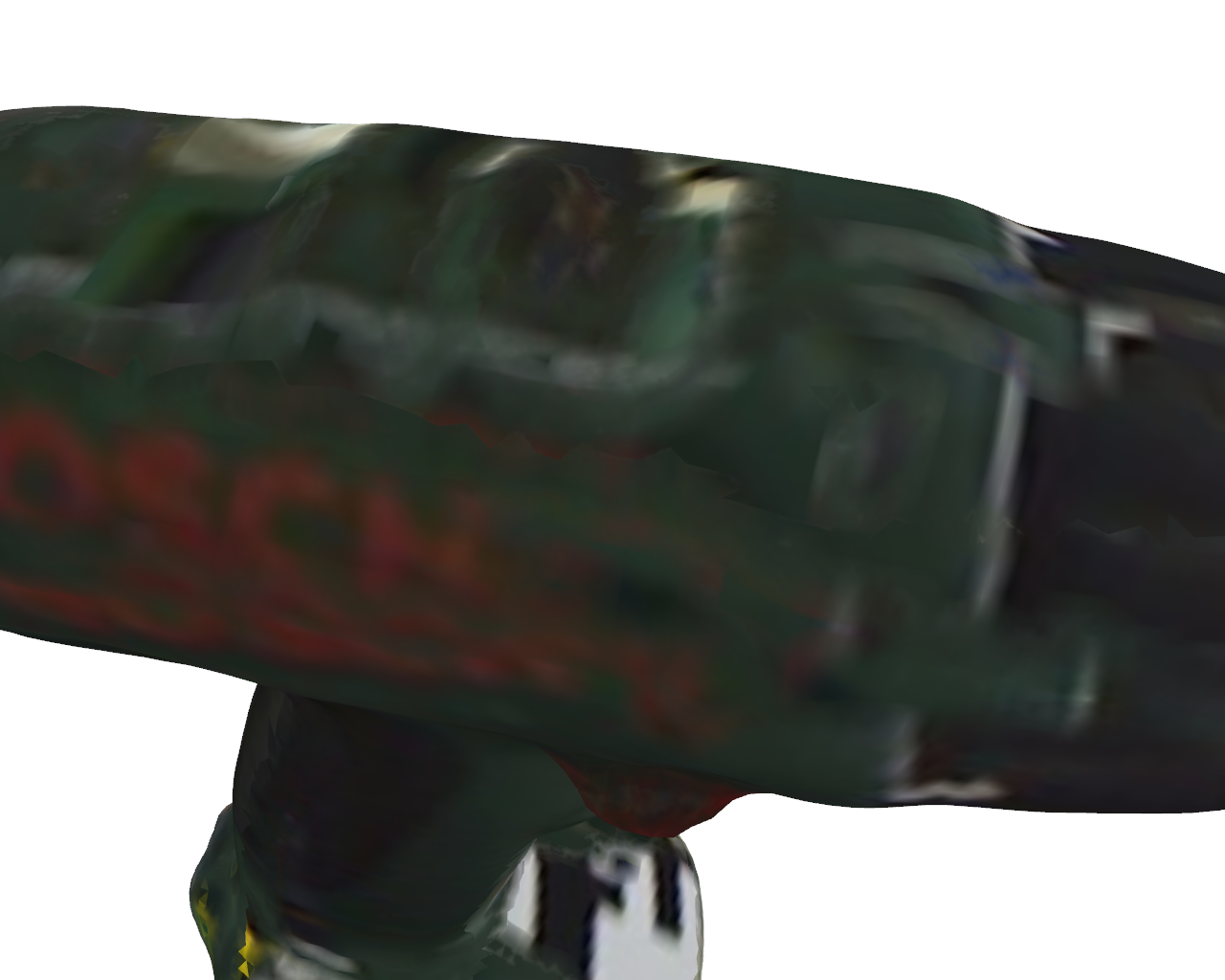}
\label{fig:texmax_lm}
}
\subfloat["Single shot pose" \cite{tekin2018real}] {
\includegraphics[width=0.3\textwidth]{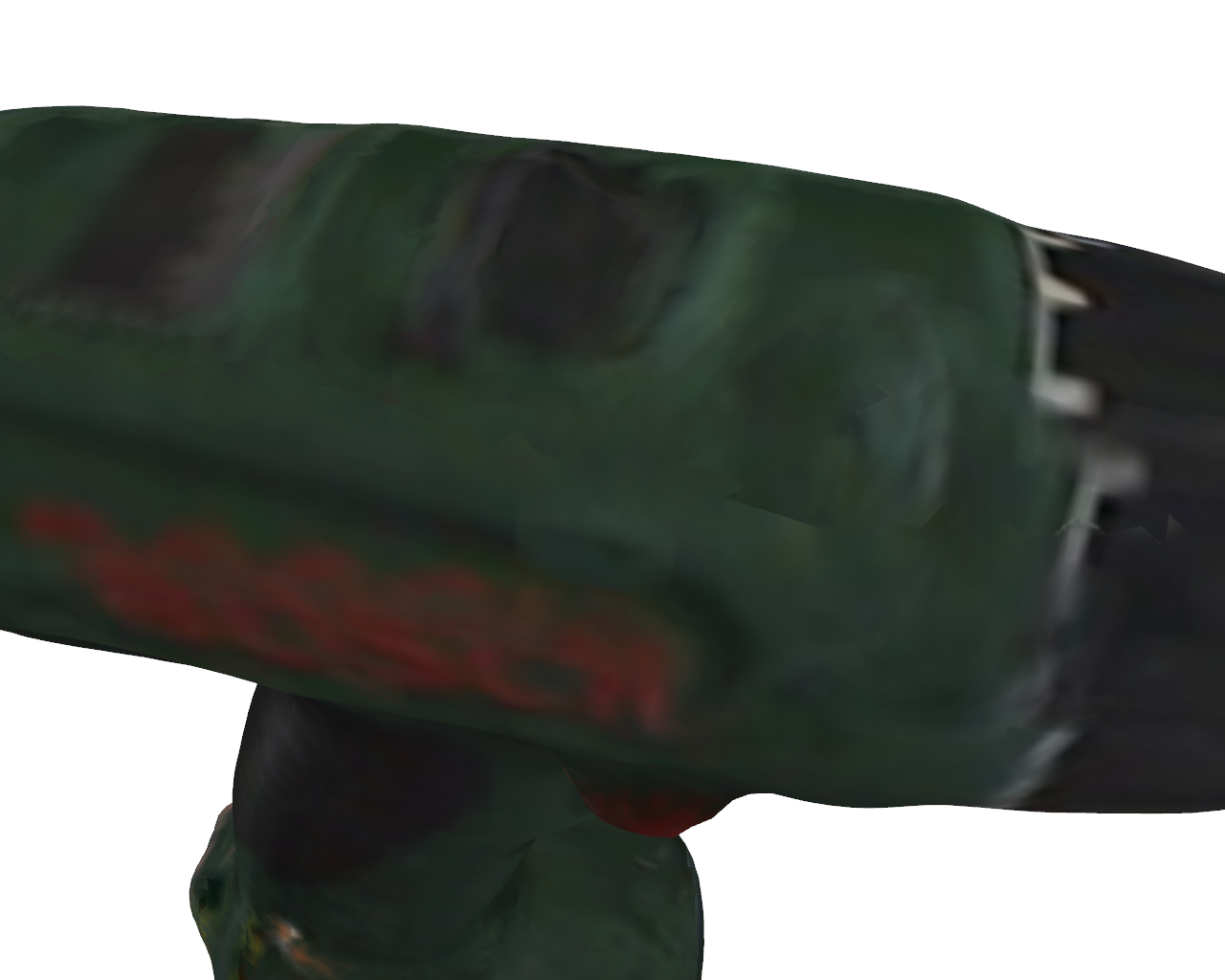}
\label{fig:texmax_ssp}
}
\subfloat[Ground truth] {
\includegraphics[width=0.3\textwidth]{figure/driller_tex.png}
\label{fig:texmax_gt}
}
\caption{Qualitative results for on-the-fly surface color reconstructions of the "driller" object in relation to different pose detection methods.}
\label{fig:on_the_fly}
\end{figure*}

The presented method is evaluated in the context of object detection. To this end we train the LINE2D variant of the LINEMOD detector on the corresponding dataset \cite{hinterstoisser2012model}. The dataset does not contain views specifically for surface reconstruction and thus represents reconstruction during detection well.  We use the publicly available LINEMOD implementation in OpenCV. 

There are 15 sequences for different objects, consisting of RGB-D frames with ground-truth poses and recorded at distances of 65cm-115cm. We select a subset of 8 objects for which a 3D mesh is available and that are large enough to provide a reasonable texture resolution.
The meshes included in the dataset were recorded using a variation of KinectFusion \cite{newcombe2011kinectfusion} and thus encode surface information as vertex-colors.

We apply our texturing algorithm on each sequence using the ground-truth poses to merely simulate a tracking algorithm for better reproducibility. Then we train LINEMOD on synthetic renderings using the generated textures as well as included vertex-colors as a baseline.
We parametrize training and testing as \cite{hinterstoisser2012model}, particularly;
\begin{itemize}
\item We use 89 views on the upper hemisphere around the object, derived by subdividing an icosahedron twice recursively.
\item For each view there are 7 in-plane rotations with roll angles between $-45^{\circ}$ and $45^{\circ}$.
\item Furthermore 6 distances, with 10cm increments, between 65cm and 115cm are used.
\item During color based outlier rejection we discard candidates where less than 70\% of the pixels have the expected color. The threshold on per-pixel hue difference is set to $54^{\circ}$.
\end{itemize}
This results in a total of 3738 templates per object for training. However, in contrast to \cite{hinterstoisser2012model}, we are only using RGB data without depth --- therefore we do not restrict the color gradient features to the contour, but compute them on the interior as well. 

For testing, we measure the true positive rate on the sequences. As in \cite{hinterstoisser2011multimodal} we consider an object successfully detected when it is within a fixed radius $r$ around the ground truth position. We globally set $r = 11$cm in our experiments to allow for depth mis-classification by one step.

For keeping interactive performance we only consider the first 30 LINEMOD candidates for matching and outlier rejection.

To simulate the CAD data use-case without any surface information available, we additionally perform training using a white diffuse material for all objects. For generating gradient features on the interior of the object, we use ambient occlusion (AO) \cite{bavoil2008screen} as a lighting approximation. Ambient occlusion is a purely geometrical method that is independent of actual light and surface properties. We skip the outlier-rejection step as no color information is available.

\begin{table}
\centering
\begin{tabular}{|c|c|c|c|c|}
\hline 
Object & AO & vertexcolor & texture \eqref{eq:max} & texture \eqref{eq:mix} \\ 
\hline 
\hline
benchvise & 0.54 & 0.75 & \textbf{0.82} & 0.82 \\
\hline
driller & 0.15 & 0.43 & \textbf{0.63} & 0.54 \\
\hline
iron & 0.53 & \textbf{0.71} & 0.67 & 0.68 \\
\hline
can & 0.38 & 0.67 & 0.78 & \textbf{0.83} \\
\hline
glue & 0.07 & \textbf{0.21} & 0.17 & 0.17 \\
\hline
cam & 0.1 & 0.28 & \textbf{0.62} & 0.55 \\
\hline
eggbox & 0.47 & 0.6 & \textbf{0.79} & 0.79 \\
\hline
holepuncher & 0.2 & 0.62 & 0.59 & \textbf{0.65} \\
\hline
\hline
average & 0.3 & 0.53 & \textbf{0.64} & 0.63 \\
\hline
\end{tabular} 
\caption{True positive rates on the linemod dataset with different training data. The ambient occlusion (AO) variant does not include any outlier rejection.}
\label{tbl:evaluation}
\end{table}

Table \ref{tbl:evaluation} shows the true positive rates for the variants mentioned above --- as can be seen the texture based variants outperform the vertex-color baseline of \cite{hinterstoisser2012model} by a margin of 10\% on average.
However, there are strong variations between the individual objects, therefore it remains inconclusive whether variant \eqref{eq:mix} or variant \eqref{eq:max} of our algorithm is preferable.

Notably the AO variant cannot reach the performance of the other methods. With some objects where it even becomes unusable (e.g. driller, cam). This emphasizes the need of surface information for object detection.


\subsection{Using noisy pose data}

To evaluate the applicability of our method for on-the-fly texturing with noisy pose data, we additionally used the state-of-the-art "single shot pose" (SSP) detector \cite{tekin2018real} instead of relying on ground-truth poses.

Figure \ref{fig:on_the_fly} shows qualitative results of texturing using ground-truth, SSP and LINEMOD poses. While the LINEMOD results only allow for for a rough color based outlier rejection, the results using SSP poses are very similar to using the ground truth.
To further quantify this, we repeated the training of SSP using synthetic renderings of the "driller" object instead of using cross-validation as in the original paper. At this, we measured the true positive rate (TPR) using the 5cm, 5deg metric.
Here, training with textured renderings (Fig. \ref{fig:texmax_gt}) resulted in a TPR of $0.37$. Using the imperfectly textured objects (Fig. \ref{fig:texmax_ssp}) resulted in a TPR of $0.34$, which supports the qualitative impression.
When training with vertex colored renderings only, the performance was significantly degraded, resulting in a TPR of $0.14$.

\subsection{Speed}

The evaluation was performed on a notebook with an Intel i7-7700HQ CPU at 2.80GHz and an Intel HD 630 iGPU. The average time to accumulate one video frame into a 1024x1024 px sized texture is 2.69 ms.
This allows running the texturing algorithm in parallel to tracking to reconstruct a texture on-the-fly.

The average time to perform a texture lookup as described in Section \ref{sec:recognition} is 0.82 ms using the software remap implementation in OpenCV.
This step can be therefore applied generally without requiring GPU usage.

\subsection{Multi instance detection}

For the multi-instance detection we performed a qualitative analysis using a separate sequence where two toy cars are alternately and simultaneously visible. The surface colors are white and red which are adjacent in HSV space (white is mapped to yellow as described in section \ref{sec:recognition}).
Furthermore, the surface exhibits specular reflection which is not filtered during texturing.

Nevertheless, our method was able to robustly discriminate both objects (see Figure \ref{fig:instances} and supplemental material\footnote{\url{https://youtu.be/IB19rTXUOt8}}).


\section{Conclusion and future work}
\label{sec:conclusion}

We have presented a method for real-time texturing that can be used to improve detection on-the-fly. At this we have shown that texturing itself is crucial for detection of CAD data where no surface information is available.
However, even for meshes where vertex-colors were previously available, our approach improves detection performance significantly.
Furthermore, we successfully applied the resulting textures to extend LINEMOD for object-instance recognition.
By interleaving detection and texture extraction it now becomes possible to extend detection algorithms by color cues on-the-fly.

Our method currently requires the camera exposure to be fixed or relies on a global exposure compensation approach which is error-prone. Here reading the actual camera exposure could be used for accurate exposure fusion of the images.
The surface specularity could be explicitly considered during merging \cite{jachnik2012real}. Currently we assume diffuse reflection, which systematically over-brightens specular surfaces. At this a plausibility test during merging could be used to reject implausible colors as caused by e.g. occlusion.
As the LINEMOD detector is no longer state-of-the-art and further investigation is needed to similarly integrate our approach into an existing CNN based method. This will require to breaking up the end-to-end trained "black-box" to make the color information explicit.

\bibliographystyle{abbrvnat}
\bibliography{bibliography}

\end{document}